\documentclass[10pt,twocolumn,letterpaper]{article}

\usepackage{cvpr}              

\usepackage[accsupp]{axessibility}
\usepackage{capt-of}
\usepackage[dvipsnames]{xcolor}
\usepackage{xspace}
\usepackage{enumitem}
\usepackage{amsmath,amssymb,amsbsy,amsfonts,dsfont,pifont,bm,bbm,mathrsfs,mathtools,nicefrac}
\usepackage{algorithm,algpseudocode,listings}
\usepackage{booktabs,multirow,adjustbox,diagbox,threeparttable,tabularray}

\definecolor{cvprblue}{rgb}{0.21,0.49,0.74}
\usepackage[pagebackref,breaklinks,colorlinks,citecolor=cvprblue]{hyperref}
\usepackage{wrapfig}
\usepackage[capitalize]{cleveref}  
\newtheorem{theorem}{Theorem}

\crefname{section}{Sec.}{Secs.}
\Crefname{section}{Section}{Sections}
\crefname{appendix}{App.}{Apps.}
\Crefname{appendix}{Appendix}{Appendices}
\crefname{table}{Tab.}{Tabs.}
\Crefname{table}{Table}{Tables}
\crefname{figure}{Fig.}{Figs.}
\Crefname{figure}{Figure}{Figures}
\crefname{equation}{Eq.}{Eqs.}
\Crefname{equation}{Equation}{Equations}
\crefname{theorem}{Thm.}{Thms.}
\Crefname{theorem}{Theorem}{Theorems}
\crefname{lemma}{Lem.}{Lems.}
\Crefname{lemma}{Lemma}{Lemmas}
\crefname{remark}{Rem.}{Rems.}
\Crefname{remark}{Remark}{Remarks}
\crefname{corollary}{Cor.}{Cors.}
\Crefname{corollary}{Corollary}{Corollaries}
\crefname{algorithm}{Alg.}{Algs.}
\Crefname{algorithm}{Algorithm}{Algorithms}
\def\paperID{7977}
\def\confName{CVPR}
\def\confYear{2024}

\definecolor{cellred}{RGB}{213, 123, 101}
\definecolor{cellgreen}{RGB}{0, 205, 0}

\newcommand{\tocite}[1]{\textcolor{red}{[TO CITE]}}
\newcommand{\method}{\textcolor{black}{\mbox{\texttt{TimeTuner}}}\xspace}
\newcommand{\methodtitle}{\textcolor{black}{\mbox{TimeTuner}}\xspace}
\newcommand{\supp}{\textit{Supplementary Material}\xspace}

\title{Towards More Accurate Diffusion Model Acceleration with A Timestep Tuner}

\author{%
    Mengfei Xia$^1$ \quad
    Yujun Shen$^2$ \quad
    Changsong Lei$^1$ \quad
    Yu Zhou$^1$ \quad
    Deli Zhao$^3$ \\[2pt]
    Ran Yi$^4$ \quad
    Wenping Wang$^5$ \quad
    Yong-Jin Liu$^{1}$\thanks{Corresponding author.} \quad \\[5pt]
    $^1$Tsinghua University \quad
    $^2$Ant Group \quad
    $^3$Alibaba Group \\[2pt]
    $^4$Shanghai Jiao Tong University \quad
    $^5$Texas A\&M University
}

\begin{document}

\maketitle

\begin{abstract}

A diffusion model, which is formulated to produce an image using thousands of denoising steps, usually suffers from a slow inference speed.
Existing acceleration algorithms simplify the sampling by skipping most steps yet exhibit considerable performance degradation.
By viewing the generation of diffusion models as a discretized integral process, we argue that the quality drop is partly caused by applying an inaccurate integral direction to a timestep interval.
To rectify this issue, we propose a \textbf{timestep tuner} that helps find a more accurate integral direction for a particular interval at the minimum cost.
Specifically, at each denoising step, we replace the original parameterization by conditioning the network on a new timestep, enforcing the sampling distribution towards the real one.
Extensive experiments show that our plug-in design can be trained efficiently and boost the inference performance of various state-of-the-art acceleration methods, especially when there are few denoising steps.
For example, when using 10 denoising steps on LSUN Bedroom dataset, we improve the FID of DDIM from 9.65 to 6.07, simply by adopting our method for a more appropriate set of timesteps.
Code is available at \href{https://github.com/THU-LYJ-Lab/time-tuner}{https://github.com/THU-LYJ-Lab/time-tuner}.

\end{abstract}

\section{Introduction}\label{sec:intro}

Diffusion probabilistic models (DPMs)~\citep{sohl2015deep,ho2020denoising,song2020score}, known simply as {diffusion models}, have recently received growing attention due to its efficacy of modeling complex data distributions~\citep{ho2020denoising, nichol2021improved, dhariwal2021diffusion}.
A DPM first defines a forward diffusion process (\textit{i.e.}, either discrete-time~\citep{ho2020denoising,song2020denoising} or continuous-time~\citep{song2020score}) by gradually adding noise to data samples, and then learns the reverse denoising process with a timestep-conditioned parameterization.
Consequently, it usually requires thousands of denoising steps to synthesize an image, which is time-consuming~\citep{ho2020denoising,song2020denoising,kong2021fast}.

To accelerate the generation process of diffusion models, a common practice is to reduce the number of inference steps.
For example, instead of a step-by-step evolution from the state of timestep 1,000 to the state of timestep 900, previous works~\citep{song2020score,ho2020denoising,bao2022analyticdpm} manage to directly link these two states with a one-time transition.
That way, it only needs to evaluate the denoising network once instead of 100 times, thus substantially saving the computational cost.

\begin{figure*}[t]
\centering
\includegraphics[width=1.0\textwidth]{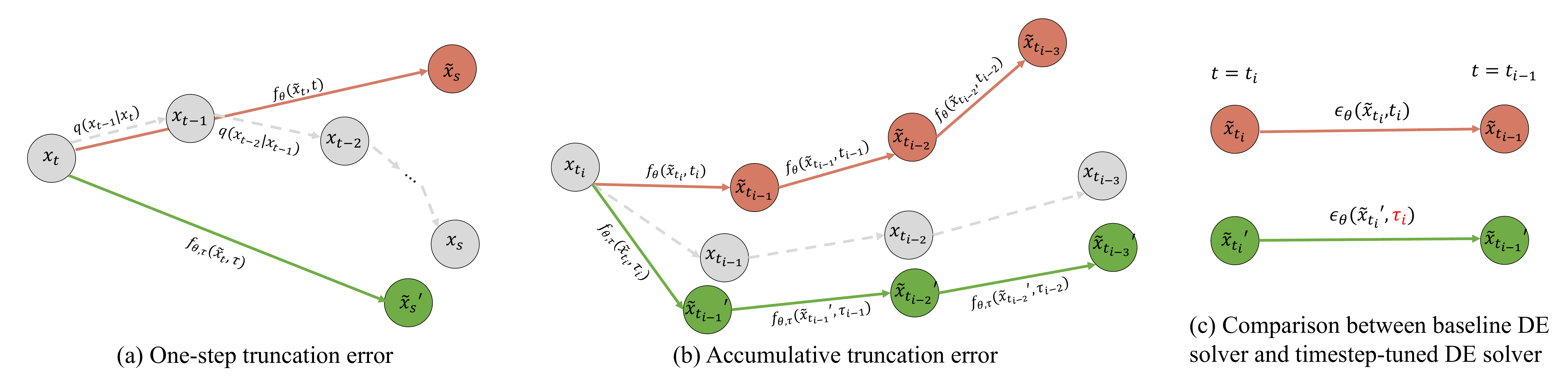}
\vspace{-15pt}
\definecolor{mygray}{RGB}{108,108,108}
\definecolor{myred}{RGB}{213, 123, 101}
\definecolor{mygreen}{RGB}{112,173,71}
\caption{%
    \textbf{Conceptual description} of (a) the one-step truncation error, (b) the accumulative truncation error, and (c) enforcing the sampling distribution towards the real distribution of our \method by replacing the input timestep from $t$ to $\tau$, by the full-step reverse process (\textbf{\textcolor{mygray}{gray dashed line}}), the baseline acceleration pipeline (\textbf{\textcolor{myred}{red line}}), and our proposed method with timestep tuner (\textbf{\textcolor{mygreen}{green line}}).
}
\label{fig:method}
\vspace{-5pt}
\end{figure*}

\definecolor{mygray}{RGB}{108, 108, 108}
\definecolor{myred}{RGB}{213, 123, 101}
\definecolor{mygreen}{RGB}{112, 173, 71}

A side effect of the above acceleration pipeline is the performance degradation that appears as artifacts in the synthesized images.
In this paper, we aim to identify and address the cause of this side effect.
As the \textcolor{mygray}{gray dashed line} shown in \cref{fig:method}a, the generation process of diffusion models can be viewed as a discretized integrating process, where the direction of each integral step is calculated by the pre-learned noise prediction model.
To reduce the number of steps, existing algorithms~\citep{song2020denoising,bao2022analyticdpm,nichol2021improved} typically apply the direction predicted for the initial state for the following timestep interval, as the \textcolor{myred}{red line} shown in \cref{fig:method}a, resulting in a gap between the sampling distribution and the real distribution.
Karras~\citep{Karras2022edm} identifies this distribution gap as the \textit{truncation error}, which accumulates over the whole steps intuitively and theoretically.
As the \textcolor{myred}{red line} shown in \cref{fig:method}b, the gap between the sampling distribution and the real distribution increases as the integrating process evolves.

To alleviate the problem arising from skipping steps, we propose a timestep tuner, termed as \method, which targets at finding a more accurate integral direction for a particular interval.
As the \textcolor{mygreen}{green line} shown in \cref{fig:method}a, our approach is designed to achieve this purpose efficiently by searching a more appropriate timestep $\tau$ replacing the previous $t$ as the new condition input of the pre-learned noise prediction model.
By doing so, we are able to significantly reduce the one-step truncation error, and hence the accumulative truncation error, as demonstrated with the \textcolor{mygreen}{green line} in \cref{fig:method}a and \cref{fig:method}b respectively.
Our motivation is intuitive -- it is based on the observation that ``\textit{although the direction estimated from the initial state may not be appropriate for integration on the interval, the one from some intermediate state can be}'', akin to the mean value theorem of integrals.
To obtain this new timestep, we enforce the sampling distribution towards the real distribution by optimizing a specially designed loss function.
We theoretically prove the feasibility of \method, and provide an estimation of the error bound for deterministic sampling algorithms.
Experiments using different numbers of function evaluations (NFE) show that our \method can be used to boost the sampling quality of various acceleration methods without extra time cost, (\textit{e.g.}, DDIM~\citep{song2020denoising}, Analytic-DPM~\citep{bao2022analyticdpm}, DPM-solver~\citep{lu2022dpm}, \textit{etc.}) in a plug-in fashion.
Hence, our work offers a new perspective on accelerating the inference while simultaneously reducing the quality degradation of diffusion models.

\section{Related Work}\label{sec:related-work}

\noindent\textbf{DPMs and the applications.} 
Diffusion probabilistic model (DPM) is initially introduced by Sohl-Dickstein~\textit{et al.}~\citep{sohl2015deep}, where the training is based on the optimization of the variational lower bound $L_{vb}$.
Denoising diffusion probabilistic model (DDPM)~\citep{ho2020denoising} proposes a re-parameterization trick of DPM and learns the reweighted $L_{vb}$.
Song~\textit{et al.}~\citep{song2020score} model the forward process as a stochastic differential equation and introduce continuous timesteps.
With rapid advances in recent studies, DPMs show great potential in various downstream applications, including speech synthesis~\citep{chen2020wavegrad,kong2020diffwave}, video synthesis~\citep{ho2022video}, super-resolution~\citep{saharia2021image,li2022srdiff}, conditional generation~\citep{choi2021ilvr}, and image-to-image translation~\citep{saharia2021palette,sasaki2021unit}.

\noindent\textbf{Faster DPMs} attempt to explore shorter trajectories rather than the complete reverse process, while ensuring the synthesis performance compared to the original DPM.
Existing methods can be divided into two categories.
The first category includes knowledge distillation~\citep{SalimansH22,luhman2021knowledge,song2023consistency}.
Although such methods may achieve respectable performance with only one-step generation~\citep{song2023consistency}, they require expensive training stages before applied to efficient sampling, leading to poor applicability.
The second category consists of training-free methods suitable for pre-trained DPMs.
DDIM~\citep{song2020denoising} is the first attempt to accelerate the sampling process using a probability flow ODE~\citep{song2020score}.
%
%
%
Some methods try to search for the trajectories by solving a least-cost-path problem with a dynamic programming (DP) algorithm or using the analytic solution~\citep{watson2021learning,bao2022analyticdpm}.
Inspired by this, seminal works conduct deeper research on trajectory choice~\citep{liu2023oms,wang2023}.
Another representative category of fast sampling methods use high-order differential equation (DE) solvers~\citep{jolicoeur2021gotta,Liu0LZ22,PopovVGSKW22,tachibana2021taylor,lu2022dpm}.
Saharia~\textit{et al.}~\citep{saharia2021image} and Ho~\textit{et al.}~\citep{ho2022cascaded} manage to train DPMs using continuous noise level and draw samples by applying a few-step discrete reverse process.
Some GAN-based methods also consider larger sampling step sizes, e.g., in \citep{xiao2022DDGAN} a multi-modal distribution is learned in a conditional GAN with a large step size.
However, to the best of our knowledge, existing training-free acceleration algorithms are bottlenecked by the poor sampling performance with extremely few inference steps (\textit{e.g.}, less than 5 steps).
Our \method can be considered as a performance booster for existing training-free acceleration methods, \textit{i.e.}, it further improves the generation performance.

\section{Method}\label{sec:method}

\subsection{Background}

Suppose that $\mathbf x_0\in\mathbb R^D$ is a $D$-dimensional random variable with an unknown distribution $q_0(\mathbf x_0)$.
DPMs~\citep{sohl2015deep,song2020score,ho2020denoising} define a forward process $\{\mathbf x_t\}_{t\in(0,T]}$ by gradually adding noise on $\mathbf x_0$ with $T>0$, such that for any $t\in(0,T]$, we have the transition distribution:
\begin{align}\label{eq:tran}
q_{0t}(\mathbf x_t|\mathbf x_0)=\mathcal N(\mathbf x_t;\alpha_t\mathbf x_0,\sigma_t^2\mathbf I),
\end{align}
where $\alpha_t,\sigma_t\in\mathbb R^+$ are differentiable functions of $t$ with bounded derivatives.
The choice of $\alpha_t,\sigma_t$ is referred to as the \textit{noise schedule}.
Let $q_t(\mathbf x_t)$ be the marginal distribution of $\mathbf x_t$, DPM ensures that $q_T(\mathbf x_T)\approx\mathcal N(\mathbf x_T;\mathbf 0,\sigma^2\mathbf I)$ for some $\sigma>0$, and the signal-to-noise-ratio (SNR) $\alpha_t^2/\sigma_t^2$ is strictly decreasing with respect to timestep $t$~\citep{kingma2021variational}.

Seminal works~\cite{kingma2021variational,song2020score} studied the underlying stochastic differential equation (SDE) theory of DPMs.
The forward and reverse processes are as below for any $t\in(0,T]$:
\vspace{-6pt}
\begin{align}
\mathrm d\mathbf x_t&=f(t)\mathbf x_t\mathrm dt+g(t)\mathrm d\mathbf w_t,\quad\mathbf x_0\sim q_0(\mathbf x_0),\label{eq:forward}\\
\mathrm d\mathbf x_t&=[f(t)\mathbf x_t-g^2(t)\nabla_{\mathbf x_t}\log q_t(\mathbf x_t)]\mathrm dt+g(t)\mathrm d\bar{\mathbf w}_t,\label{eq:reverse}
\end{align}
where $\mathbf w_t,\bar{\mathbf w}_t$ are standard Wiener processes in forward and reverse time, respectively, and $f,g$ have closed-form expressions w.r.t $\alpha_t,\sigma_t$.
The unknown $\nabla_{\mathbf x_t}\log q_t(\mathbf x_t)$ is referred to as the \textit{score function}.
Furthermore, Song~\textit{et al.}~\citep{song2020denoising} propose the \textit{probability flow ODE} for faster and more stable sampling, which enjoys the identical marginal distribution at each $t$ as that of the SDE in \cref{eq:reverse}, \textit{i.e.},
\vspace{-6pt}
\begin{align}\label{eq:probODE}
\frac{\mathrm d\mathbf x_t}{\mathrm dt}=f(t)\mathbf x_t-\frac{1}{2}g^2(t)\nabla_{\mathbf x_t}\log q_t(\mathbf x_t).
\end{align}

Technically, DPMs implement sampling by solving DEs with numerical solvers discretizing the DE from $T$ to 0.
To this end, DPMs introduce a neural network $\boldsymbol \epsilon_\theta(\mathbf x_t,t)$, namely the noise prediction model, to approximate the score function from the given $\mathbf x_t$, where the parameter $\theta$ can be optimized by the objective below:
\begin{align}\label{eq:dpm_loss}
\mathbb E_{\mathbf x_0,\epsilon,t}\left[\omega_t\|\boldsymbol \epsilon_\theta(\mathbf x_t,t) - \boldsymbol\epsilon\|_2^2\right],
\end{align}
where $\omega_t$ is the weighting function, $\boldsymbol\epsilon\sim\mathcal N(\mathbf 0,\mathbf I)$, $\mathbf x_t=\alpha_t\mathbf x_0+\sigma_t\boldsymbol\epsilon$, and $t\sim\mathcal U[0,T]$.


\definecolor{myblue}{RGB}{132, 180, 213}
\definecolor{myred}{RGB}{213, 123, 101}
\begin{figure*}[t]
\centering
\includegraphics[width=0.95\textwidth]{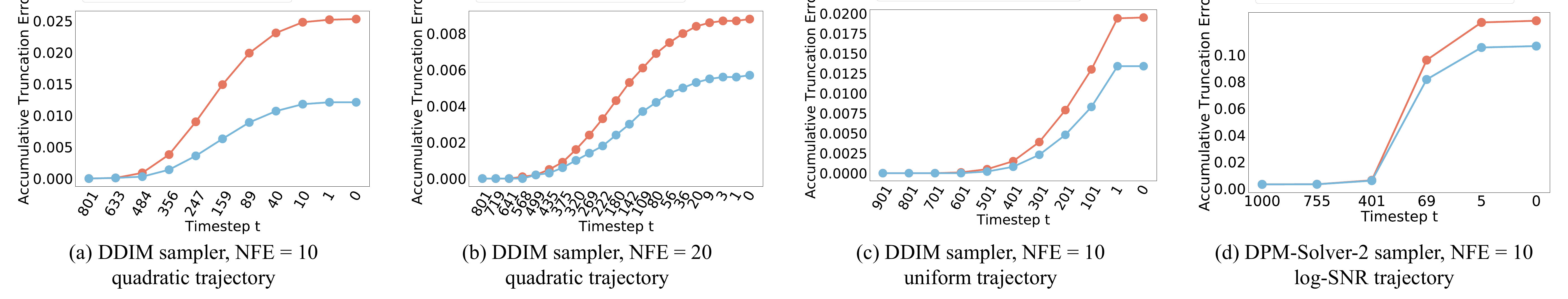}
\vspace{-10pt}
\caption{%
    \textbf{Quantitative measurement} of the gap between real and sampling distribution using DDIM and DPM-Solver-2.
    The horizontal axis represents timesteps forming (a) quadratic trajectory with NFE $=10$; (b) quadratic trajectory with NFE $=20$; (c) uniform trajectory with NFE $=10$; (d) log-SNR trajectory with NFE $=10$.
    We plot the $L_2$ distance between $(\mathbf x_t,\widetilde{\mathbf x}_t)$ for the original and the timestep-tuned sampler, shown in \textbf{\textcolor{myred}{red}} and \textbf{\textcolor{myblue}{blue}}, respectively.
    We also provide an error bound for deterministic sampler theoretically in \Cref{thm:main2}.
}
\label{fig:gap}
\vspace{-9pt}
\end{figure*}

\subsection{Gap between Real and Sampling Distributions}\label{subsec:3.gap}

Recall that a DPM sampler is built on the noise prediction model $\boldsymbol\epsilon_\theta$ at each timestep $t$, which is a discretization of an integrating process.
At each denoising step, one applies $\boldsymbol\epsilon_\theta$ on the intermediate result $\widetilde{\mathbf x}_t$ from the last denoising step together with its corresponding timestep condition $t$, \textit{i.e.}, $\widetilde{\boldsymbol\epsilon}_t=\boldsymbol\epsilon_\theta(\widetilde{\mathbf x}_t,t)$.
The predicted noise $\widetilde{\boldsymbol\epsilon}_t$ will be used as the integral direction towards the next denoised result.

However, recall that during the training of DPM, the noise prediction model $\boldsymbol\epsilon_\theta$ is trained with the noisy data at timestep $t$, \textit{i.e.}, $\mathbf x_t=\alpha_t\mathbf x_0+\sigma_t\boldsymbol\epsilon$, where $\boldsymbol\epsilon\sim\mathcal N(\mathbf 0,\mathbf I)$.
Intuitively, due to the error of the DE solver (\textit{i.e.}, SDE or ODE solver) using large discretization step sizes, there is a considerable gap between the real distribution of $\mathbf x_t$ and the sampling distribution of $\widetilde{\mathbf x}_t$ at each timestep $t$, which is called the truncation error~\citep{Karras2022edm}.
Even worse, this error is accumulated progressively during the reverse process,
since the gap between the distributions of $\mathbf x_t$ and $\widetilde{\mathbf x}_t$ leads to a poor prediction of $\widetilde{\mathbf x}_{t-1}$, in which the truncation error at timestep $t$ transmits to the next timestep $t-1$~\citep{Karras2022edm}.

To support this insight, \cref{fig:method}a shows the truncation error at each timestep, in comparison with the original full-step sampling process.
As the \textcolor{mygray}{gray dashed line} in \cref{fig:method}a indicates, in the original full-step sampling process, the denoised $\mathbf x_s$ comes from a step-by-step denoising refinement starting at $\mathbf x_t$, by the results from $\boldsymbol\epsilon_\theta$ at all intermediate timesteps from $t$ down to $s$.
However, the accelerated sampling method (\textit{i.e.}, \textcolor{myred}{red line}) uses a large step size and replaces all intermediate predicted noises with one single predicted noise in the initial timestep, \textit{i.e.}, $\widetilde{\boldsymbol\epsilon}_t = \boldsymbol\epsilon_\theta(\widetilde{\mathbf x}_t, t)$.
Therefore, each $\widetilde{\mathbf x}_t$ incurs a truncation error consequently.

Furthermore, note that for a sampling process, given $\widetilde{\mathbf x}_T$ at timestep $T$ and a timestep $t\in[0, T)$, the DE solver approximates the true $\mathbf x_t$ as $\widetilde{\mathbf x}_t$.
As the \textcolor{myred}{red line} indicates in \cref{fig:method}b, since the local truncation error accumulates at each step, the gap between the sampling distribution of $\widetilde{\mathbf x}_t$ and the real distribution of $\mathbf x_t$ increases as $t$ evolves.
To gain the insight of the accumulative error, we conduct a simple experiment to provide convincing evidence of the above observation using DDIM~\citep{song2020denoising} and DPM-Solver-2~\citep{lu2022dpm} on CIFAR10 dataset~\citep{Krizhevsky_2009_17719}.
We first sample $\widetilde{\mathbf x}_T$ and estimate $\mathbf x_t$ sequences using DDIM sampler with NFE $=1,000$.
Meanwhile, we draw the approximate $\widetilde{\mathbf x}_t$ sequences using DDIM and DPM-Solver-2 under quadratic, uniform, and log-SNR trajectories with NFE $=10$ or $20$, respectively.
Then we calculate the $L_2$ metric of $\mathbf x_t-\widetilde{\mathbf x}_t$ at each timestep $t$, in order to demonstrate the gap between the two distributions, the result is shown in \cref{fig:gap}.
It is noteworthy that: (1) the gap between $\mathbf x_t$ and $\widetilde{\mathbf x}_t$ indeed accumulates as timestep $t$ evolves from $T$ to $0$, making the sampling distribution farther and farther away from the real one, which severely hurts the final sampling quality;
(2) with the same quadratic trajectory, the larger the NFE is, the smaller the gap between the two distributions is;
(3) different types of trajectories and DPM samplers account for different behaviors of the accumulative truncation error.

\subsection{Theoretical Foundations of \methodtitle}\label{sec:3.4}

Recall that in \cref{subsec:3.gap} we demonstrate the gap between the real and the sampling distributions.
This gap will damage the quality of the synthesis samples.
In this section, we will delve into the theory of reverse denoising process in DPMs, and give an upper bound of the distribution gap for the deterministic sampler, which derives the feasibility of the \method.
Before stating the main theorem, we first propose several definitions for simplicity.
For a discretization $0=t_0<t_1<\cdots<t_K=T$ of $[0,T]$ and a DE solver denoted by $f_\theta(\mathbf x_{t_i},t_i)$, which is responsible to denoise the intermediate $\widetilde{\mathbf x}_{t_i}$ for one single step, \textit{i.e.}, $\widetilde{\mathbf x}_{t_{i-1}}=f_\theta(\widetilde{\mathbf x}_{t_i},t_i)$ for $i=1,2,\cdots,K$.
For instance, the DE solver $f_\theta$ of DDIM~\citep{song2020denoising} is of the following form:
\begin{align}
f_\theta(\widetilde{\mathbf x}_{t_i},t_i)&=\frac{\alpha_{t_{i-1}}}{\alpha_{t_i}}\widetilde{\mathbf x}_{t_i}\nonumber\\
&\qquad-\left(\frac{\alpha_{t_{i-1}}\sigma_{t_i}}{\alpha_{t_i}}-\sigma_{t_{i-1}}\right)\boldsymbol\epsilon_\theta(\widetilde{\mathbf x}_{t_i}, t_i).
\end{align}
One can generalize the definition of $f_\theta(\mathbf x_{t_i},t_i)$ as below:
\vspace{-12pt}
\begin{align}\label{eq:aligned_de}
f_{\theta,\tau}(\widetilde{\mathbf x}_{t_i},\tau_i)\nonumber&:=\frac{\alpha_{t_{i-1}}}{\alpha_{t_i}}\widetilde{\mathbf x}_{t_i}\nonumber\\
&\qquad-\left(\frac{\alpha_{t_{i-1}}\sigma_{t_i}}{\alpha_{t_i}}-\sigma_{t_{i-1}}\right)\boldsymbol\epsilon_\theta(\widetilde{\mathbf x}_{t_i}, \textcolor{red}{\tau_i}),
\end{align}
where $\tau_i$ replaces the previous input condition timestep $t_i$ for each $i=1,2,\cdots,K$.
Now we state the theorem below.
Proof is available in \supp.

\begin{theorem}\label{thm:main2}
Assume that $\boldsymbol\epsilon_\theta$ is the ground-truth noise prediction model, with $\|\boldsymbol\epsilon_\theta(\mathbf x,t)-\boldsymbol\epsilon_\theta(\mathbf y,t)\|_2\geqslant\frac{1}{C}\|\mathbf x-\mathbf y\|_2$ for any $t$ and some $C>0$.
Denote by $\mathbf x_{t_i}^{gt}$ the ground-truth intermediate result at $t_i$ starting from $\widetilde{\mathbf x}_{t_K}$, and by $f_{\theta,\tau}$ a deterministic sampler.
We have the following inequality:
\vspace{-18pt}
\begin{align}\label{eq:main}
&\mathbb E_{\mathbf x_0,\boldsymbol\epsilon}[\|\widetilde{\mathbf x}_{t_{i-1}}-\mathbf x_{t_{i-1}}^{gt}\|_2]\nonumber\\
\leqslant&C\Big(\sum_{n=i}^K\left(\mathbb E_{\mathbf x_0,\boldsymbol\epsilon}\left[\|\boldsymbol\epsilon_\theta(f_{\theta,\tau}(\widetilde{\mathbf x}_{t_i},\tau_i),t_{i-1})-\boldsymbol\epsilon_\theta(\widetilde{\mathbf x}_{t_i},t_i)\|_2^2\right]\right)^{\frac{1}{2}}\nonumber\\
&\qquad+\sum_{l=i}^K\mathbb E[\|\boldsymbol\epsilon_\theta(\mathbf x_{t_l}^{gt},t_l)-\boldsymbol\epsilon_\theta(\mathbf x_{t_{l-1}}^{gt},t_{l-1})\|_2]\Big).
\end{align}
\end{theorem}

In conclusion, \Cref{thm:main2} studies the relation between the distribution gap and the generalized DE sampler $f_{\theta,\tau}$.
Intuitively, this is based on the observation that some intermediate state could be more appropriate for the integration on the interval than the initial one, akin to the mean value theorem of integrals.
Therefore, what we need to do to boost any given acceleration algorithm, is to choose an adequate timestep $\tau$, \textit{replacing the input for $\boldsymbol\epsilon_\theta$ from $t$ to $\tau$} (which is shown in \cref{fig:method}c), such that the sampling distribution of $\widetilde{\mathbf x}_t$ tends to obey $q_t(\mathbf x_t)$ by shrinking the upper bound in \cref{eq:main}.
This profound conclusion facilitates DPM acceleration from a novel perspective.

\subsection{Timestep Tuner for Noise Prediction Model}

Now we formally propose \method, targeting more accurate DPM acceleration.
Recall that in \cref{sec:3.4} we give an analysis on distribution gap, which is bounded by a term containing generalized DE solver $f_{\theta,\tau}$ with replaced timesteps.
This motivates us to bridge the distribution gap with more proper input condition timesteps.

First we provide a picture of our formulation, as shown in \cref{fig:method}.
To reduce the truncation error caused by inaccurate integral direction and large discretization step size (\textit{i.e.}, the denoised result $\widetilde{\mathbf x}_s$ achieved along the \textcolor{myred}{red line} in \cref{fig:method}a), we target at finding a more accurate integral direction (\textit{i.e.}, by replacing $t_i$ with optimized \textcolor{red}{$\tau_i$} in \cref{fig:method}c) for the interval from $t$ to $s$ (\textit{i.e.}, the denoised result $\widetilde{\mathbf x}_s'$ achieved along the \textcolor{mygreen}{green line} in \cref{fig:method}a).
In this sense, we are able to achieve a better integral direction for each interval (\textit{i.e.}, the \textcolor{mygreen}{green line} in \cref{fig:method}b), mitigating the accumulation of the truncation error (\textit{i.e.}, the \textcolor{myred}{red line} in \cref{fig:method}b).
Formally, for a discretization $0=t_0<t_1<\cdots<t_K=T$ and a DE solver $f_\theta$, \method will employ the generalized DE solver $f_{\theta, \tau}$ to enforce $\widetilde{\mathbf x}_{t_i}$ towards $q_{t_i}(\mathbf x_{t_i})$, where $\tau_i$ is the optimized timestep replacing the previous input condition timestep $t_i$ for each $i=1,2,\cdots,K$.

\begin{algorithm}[t]
\caption{Training $\tau_i$ with sequential strategy}\label{alg:train}
\begin{algorithmic}[1]
\Repeat
\State $\mathbf x_0\sim q_0(\mathbf x_0), \boldsymbol\epsilon\sim\mathcal N(\mathbf 0,\mathbf I)$
\State $\widetilde{\mathbf x}_T\leftarrow\alpha_T\mathbf x_0+\sigma_T\boldsymbol\epsilon$
\State $\widetilde{\mathbf x}_{t_i}\leftarrow f_{\theta,\tau}(\cdots(f_{\theta,\tau}(\widetilde{\mathbf x}_T,\tau_K)\cdots),\tau_{i+1})$
\State Take gradient descent step on
$$\nabla_{\tau_i}(\|\boldsymbol\epsilon_\theta(f_{\theta,\tau}(\widetilde{\mathbf x}_{t_i},\tau_i),t_{i-1})-\boldsymbol\epsilon_\theta(\widetilde{\mathbf x}_{t_i},t_i)\|_2^2)$$
\Until converged
\end{algorithmic}
\end{algorithm}

Training \method is simple and efficient.
Recall that we hope to find a timestep $\tau_i$ replacing the previous $t_i$, such that the sampling distribution of $\widetilde{\mathbf x}_{t_{i-1}}=f_{\theta,\tau}(\widetilde{\mathbf x}_{t_i},\tau_i)$ tends to follow the real distribution $q_{t_{i-1}}(\mathbf x_{t_{i-1}})$ for each $i=1,2,\cdots,K$.
Therefore, after determining the number of function evaluations (NFE) $K$ and its corresponding trajectory $0=t_0<t_1<\cdots<t_K=T$, the loss function of $\tau_i$ is defined as below for $i=K,K-1,\cdots,1$:
\vspace{-6pt}
\begin{align}\label{eq:loss}
&\mathcal L_i(\tau_i)\nonumber\\
=&\mathbb E_{\mathbf x_0,\boldsymbol\epsilon}\left[\|\boldsymbol\epsilon_\theta(f_{\theta,\tau}(\widetilde{\mathbf x}_{t_i},\tau_i),t_{i-1})-\boldsymbol\epsilon_\theta(\widetilde{\mathbf x}_{t_i},t_i)\|_2^2\right].
\end{align}
Then according to \Cref{thm:main2}, optimizing $\mathcal L_i$ will narrow the gap between sampling and real distributions.
Therefore, we can optimize the timesteps reversely from $t_K=T$ down to $t_1$, in which $\widetilde{\mathbf x}_{t_K}=\widetilde{\mathbf x}_T=\alpha_T\mathbf x_0+\sigma_T\boldsymbol\epsilon\approx\mathcal N(\mathbf 0,\sigma^2\mathbf I)$ for given $\mathbf x_0\sim q_0(\mathbf x_0)$ and $\boldsymbol\epsilon\sim\mathcal N(\mathbf 0,\mathbf I)$.
The training process is summarized in \Cref{alg:train}.
It is noteworthy that the training of our method is in a plug-in fashion.
There is no need to modify the parameters of the pre-trained DPMs.
Besides, since it only requires optimization of a scalar $\tau_i$, the training is extremely efficient (\textit{e.g.}, \method takes $\sim1$ hour in all for NFE $=10$ on an NVIDIA A100 GPU).

\begin{figure*}[t]
\centering
\includegraphics[width=0.95\textwidth]{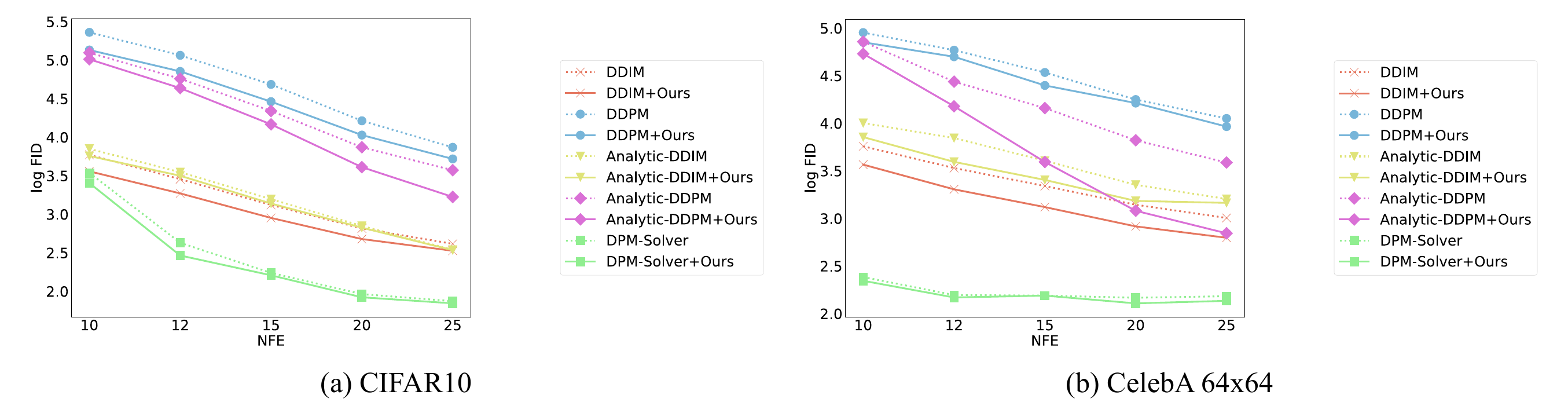}
\vspace{-10pt}
\caption{%
    \textbf{Quantitative comparison} measured by $\log$ FID $\downarrow$ on CIFAR10 and CelebA, under original DDPM.
    All are evaluated with different NFEs on the horizontal axis.
    We apply quadratic trajectory for DDIM and DDPM, uniform trajectory for Analytic-DDIM and Analytic-DDPM, log-SNR trajectory for DPM-Solver-2.
}
\label{fig:fid}
\vspace{-10pt}
\end{figure*}

\begin{figure*}[t]
\centering
\includegraphics[width=0.9\textwidth]{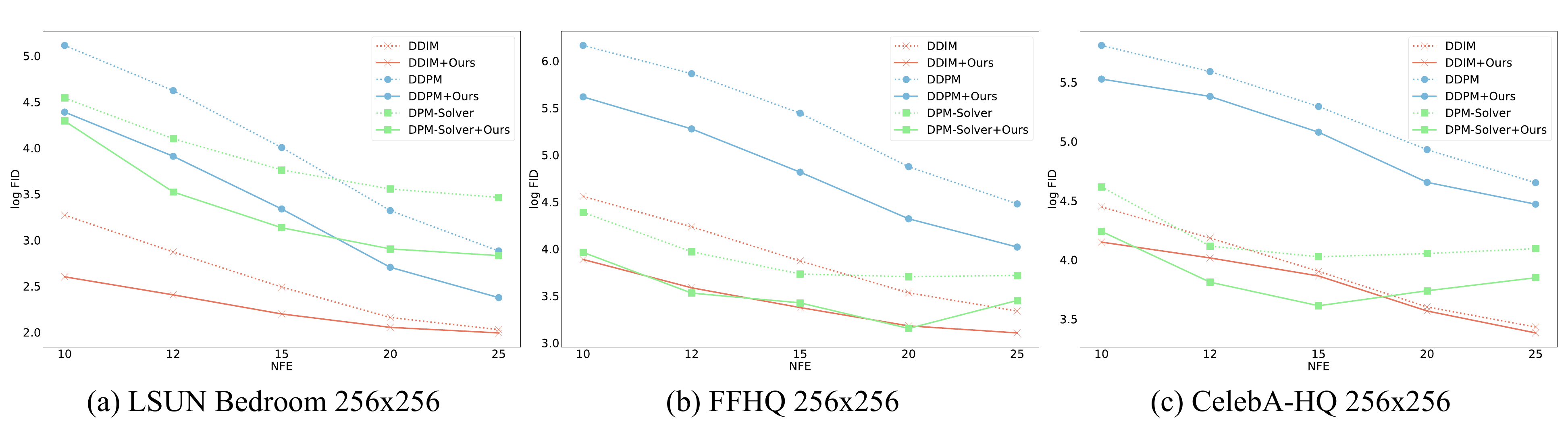}
\vspace{-10pt}
\caption{%
    \textbf{Quantitative comparison} measured by $\log$ FID $\downarrow$ on LSUN Bedroom, FFHQ, and CelebA-HQ, under LDM.
    All are evaluated with different NFEs on the horizontal axis.
    We apply uniform trajectory for DDIM and DDPM, and log-SNR trajectory for DPM-Solver-2.
}
\label{fig:fid_ldm}
\vspace{-10pt}
\end{figure*}

\begin{figure*}[!htp]
\centering
\includegraphics[width=0.85\textwidth]{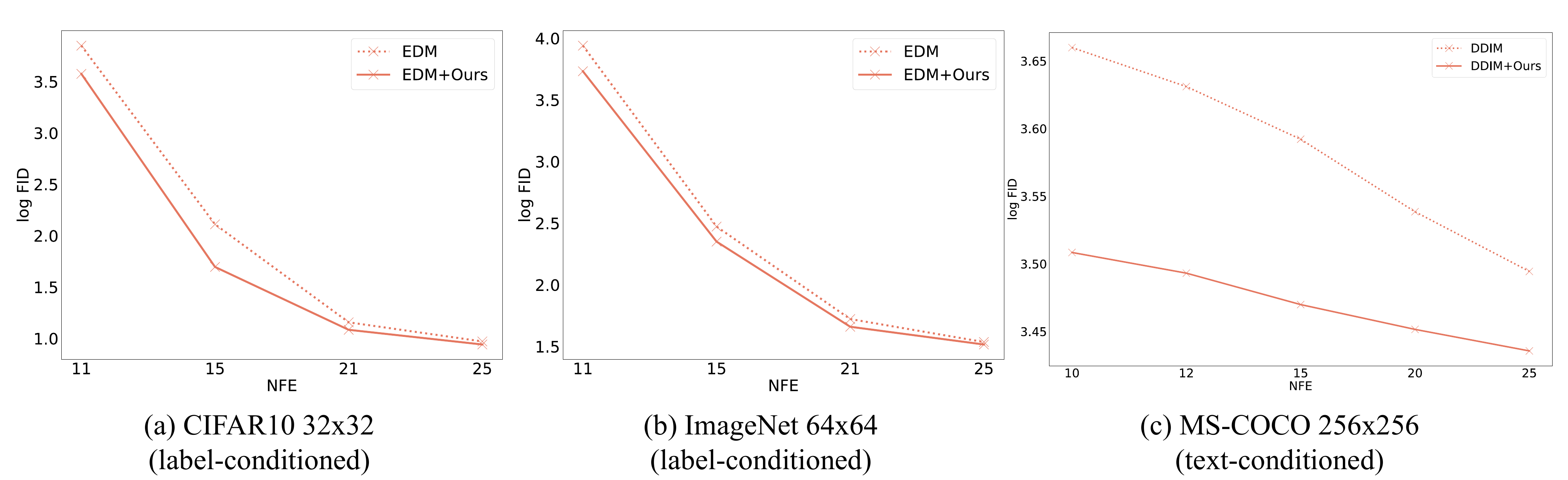}
\vspace{-10pt}
\caption{%
    \textbf{Quantitative comparison} measured by $\log$ FID $\downarrow$ on CIFAR10, ImageNet, and MS-COCO, under EDM and LDM.
    All are evaluated with different NFEs on the horizontal axis.
    We apply the originally designed trajectory for EDM and linear trajectory for LDM.
}
\label{fig:fid_cond}
\vspace{-5pt}
\end{figure*}

We then exhibit the relation between the objective of our method (\textit{i.e.}, \cref{eq:loss}) and that of the DDPM (\textit{i.e.}, \cref{eq:dpm_loss}), concluded as the theorem below.
Proof and detailed quantitative comparison are addressed in \supp.

\begin{theorem}\label{thm:main}
Assume that $\boldsymbol\epsilon_\theta$ is the ground-truth noise prediction model.
The loss function of \method resembles that of the original DPM, \textit{i.e.}, for $i=K,K-1,\cdots,1$, the optimal $\tau_i$ holds the following property:
\vspace{-8pt}
\begin{align}
&\mathop{\arg\min}_{\tau_i}\mathcal L_i(\tau_i)\nonumber\\
=&\mathop{\arg\min}_{\tau_i}\mathbb E_{\mathbf x_0,\boldsymbol\epsilon}[\|\boldsymbol\epsilon_\theta(f_{\theta,\tau}(\widetilde{\mathbf x}_{t_i},\tau_i),t_{i-1})\nonumber\\
&\qquad\qquad\qquad\qquad\qquad-\frac{\widetilde{\mathbf x}_{t_i}-\alpha_{t_i}\mathbf x_0}{\sigma_{t_i}}\|_2^2].
\end{align}
\end{theorem}

\setlength{\tabcolsep}{16pt}
\begin{table*}[t]
\caption{%
    \textbf{Quantitative comparison} measured by IS~$\uparrow$, FID~$\downarrow$, sFID~$\downarrow$, Precision~$\uparrow$ and Recall~$\uparrow$ on LSUN Bedroom, FFHQ, CelebA-HQ, and ImageNet.
    All are evaluated by drawing 50,000 samples via DDIM upon LDM, with NFE $= 10$.
    }
\label{tab:metric}
\vspace{-10pt}
\centering
\SetTblrInner{rowsep=0.3pt}       
\SetTblrInner{colsep=16.0pt}      
\begin{tblr}{
    cell{2-4,6-8,10-12,14-16}{1}={halign=l,valign=m},   
    cell{1,5,9,13}{1-6}={halign=c,valign=m},     
    cell{2-4,6-8,10-12,14-16}{2-6}={halign=c,valign=m},     
    cell{1,5,9,13}{1}={c=6}{},          
    hline{1,2,5,6,9,10,13,14,17}={1-6}{1.0pt},     
    hline{3,7,11,15}={1-6}{},          
}
LSUN Bedroom 256x256, \textit{unconditional} generation &&&&& \\
Method & IS~$\uparrow$ & FID~$\downarrow$ & sFID~$\downarrow$ & Precision~$\uparrow$ & Recall~$\uparrow$  \\
DDIM & 2.30 & 9.46 & 12.02 & 0.55 & 0.34  \\
DDIM + Ours & \bf 2.31 & \bf 5.85 & \bf 9.44 & \bf 0.57 & \bf 0.44 \\
FFHQ 256x256, \textit{unconditional} generation &&&&& \\
Method & IS~$\uparrow$ & FID~$\downarrow$ & sFID~$\downarrow$ & Precision~$\uparrow$ & Recall~$\uparrow$  \\
DDIM & 4.00 & 23.58 & 14.59 & 0.63 & 0.21 \\
DDIM + Ours & \bf 4.40 & \bf 14.80 & \bf 9.69 & \bf 0.67 & \bf 0.32 \\
CelebA-HQ 256x256, \textit{unconditional} generation &&&&& \\
Method & IS~$\uparrow$ & FID~$\downarrow$ & sFID~$\downarrow$ & Precision~$\uparrow$ & Recall~$\uparrow$  \\
DDIM & 2.95 & 18.72 & 16.68 & \bf 0.68 & 0.19 \\
DDIM + Ours & \bf 3.20 & \bf 16.59 & \bf 15.61 & 0.67 & \bf 0.26 \\
ImageNet 256x256, \textit{conditional} generation &&&&&\\
Method & IS~$\uparrow$ & FID~$\downarrow$ & sFID~$\downarrow$ & Precision~$\uparrow$ & Recall~$\uparrow$  \\
DDIM & 324.52 & 10.13 & 12.52 & 0.91 & 0.28  \\
DDIM + Ours & \bf 336.94 & \bf 9.63 & \bf 7.29 & \bf 0.92 & \bf 0.30 \\
\end{tblr}
\vspace{-0pt}
\end{table*}

\setlength{\tabcolsep}{6pt}
\begin{table*}[t]
\caption{%
    \textbf{Quantitative comparison} measured by FID~$\downarrow$, Precision~$\uparrow$, and Recall~$\uparrow$ on LSUN Bedroom by drawing 50,000 samples via CD.
    }
\label{tab:metric_cd}
\vspace{-10pt}
\centering
\SetTblrInner{rowsep=0.3pt}      
\SetTblrInner{colsep=6.0pt}      
\begin{tblr}{
    cell{1-3}{1,5}={halign=l,valign=m},   
    cell{1-3}{2-4,6-8}={halign=c,valign=m},     
    hline{1,4}={1-8}{1.0pt},     
    hline{2}={1-8}{},            
    vline{5}={1-3}{},            
}
Method & FID~$\downarrow$ & Precision~$\uparrow$ & Recall~$\uparrow$ & Method & FID~$\downarrow$ & Precision~$\uparrow$ & Recall~$\uparrow$ \\
CD (NFE $=1$) & 8.14 & \bf 0.68 & 0.32 & CD (NFE $=2$) & 5.91 & \bf 0.69 & 0.38 \\
CD + Ours (NFE $=1$) & \bf 7.14 & 0.65 & \bf 0.35 & CD + Ours (NFE $=2$) & \bf 5.18 & 0.68 & \bf 0.39 \\
\end{tblr}
\vspace{-0pt}
\end{table*}
\setlength{\tabcolsep}{6pt}
\begin{table*}[!htp]
\caption{%
    \textbf{Quantitative comparison} measured by FID~$\downarrow$, Precision~$\uparrow$, and Recall~$\uparrow$ on ImageNet and MS-COCO under LDM.
    All are evaluated via DDIM with 10 NFEs, using different CFG scales.
    For clearer demonstration, improvements are highlighted in \textcolor{cellgreen}{\textbf{green}}.
    }
\label{tab:cfg}
\vspace{-10pt}
\centering
\SetTblrInner{rowsep=0.3pt}       
\SetTblrInner{colsep=10.0pt}      
\begin{tblr}{
    cells={halign=c,valign=m},   
    cell{1}{1}={r=2}{},          
    cell{1}{2}={c=3}{},          
    hline{1,6}={1-5}{1.0pt},     
    hline{3,4,5}={1-5}{},        
    hline{2}={2-5}{},            
    vline{2,5}={1-5}{},          
}
DDIM, NFE $=$ 10 &                ImageNet 256x256 &                                &                                &                MS-COCO 256x256         \\
                 &                FID~$\downarrow$ &           Precision~$\uparrow$ &              Recall~$\uparrow$ &                FID~$\downarrow$ \\
CFG scale $=$ 3  & 10.13 \color{cellgreen} (-0.50) & 0.91 \color{cellgreen} (+0.01) & 0.28 \color{cellgreen} (+0.02) & 11.00 \color{cellgreen} (-0.47) \\
CFG scale $=$ 5  & 16.78 \color{cellgreen} (-0.39) &                    0.94 (0.00) &                    0.16 (0.00) & 12.64 \color{cellgreen} (-1.26) \\
CFG scale $=$ 7  & 20.32 \color{cellgreen} (-0.80) & 0.94 \color{cellgreen} (+0.04) &                    0.11 (0.00) & 14.80 \color{cellgreen} (-0.64) \\
\end{tblr}
\vspace{-0pt}
\end{table*}

\begin{table*}[!htp]
\caption{%
    \textbf{Quantitative comparison} measured by FID~$\downarrow$ on CIFAR10.
    All are evaluated by drawing 50,000 samples via DDIM sampler.
    }
\label{tab:comparison}
\vspace{-10pt}
\centering
\SetTblrInner{rowsep=0.3pt}       
\SetTblrInner{colsep=10.0pt}      
\begin{tblr}{
    cells={halign=c,valign=m},   
    cell{1}{1}={r=2}{},          
    cell{1}{3,6}={c=3}{},        
    hline{1,4}={1-8}{1.0pt},     
    hline{2}={2-8}{},            
    hline{3}={1-8}{},            
    vline{2,3,6}={1-3}{},        
}
Method           & NFE $=$ 16 & NFE $=$ 10, linear trajectory &          &           & NFE $=$ 10, quadratic trajectory &         &           \\
                 & GGDM~\cite{watson2021learning} & DDIM               & RS-DDIM~\cite{wang2023} &      Ours & DDIM             & OMS-DPM~\cite{liu2023oms} &      Ours \\
FID~$\downarrow$ &     $>$ 40 &              16.68 &    16.12 & \bf 15.76 &            13.65 &   12.34 & \bf 11.77 \\
\end{tblr}
\vspace{-0pt}
\end{table*}

\subsection{Parallel Training Strategy of \methodtitle}\label{subsec:3.strategy}

Recall that we employ a \textit{sequential strategy} to train each $\tau_i$ for $i$ from $K$ to $1$, \textit{i.e.}, each $\tau_i$ needs to be optimized sequentially.
Also, one can train all $\tau_i$'s simultaneously with \cref{eq:loss_parallel} by a \textit{parallel strategy} for $i=1,2,\cdots,K$:
\vspace{-18pt}
\begin{align}\label{eq:loss_parallel}
&\mathcal L_i^{parallel}(\tau_i)\nonumber\\
=&\mathbb E_{\mathbf x_0,\boldsymbol\epsilon}\left[\|\boldsymbol\epsilon_\theta(f_{\theta,\tau}(\mathbf x_{t_i},\tau_i),t_{i-1})-\boldsymbol\epsilon_\theta(\mathbf x_{t_i},t_i)\|_2^2\right],
\end{align}
where $\mathbf x_{t_i}=\alpha_{t_i}\mathbf x_0+\sigma_{t_i}\boldsymbol\epsilon$ instead of the intermediate result $\widetilde{\mathbf x}_{t_i}$ by $f_{\theta,\tau}$.
The parallel strategy is feasible since $\mathbf x_{t_i}$ is independent with all $\tau_i$'s, and one can train $\tau_i$ simultaneously by sampling different $\mathbf x_{t_i}$ on different GPUs.
Despite the extra acceleration of the optimization process, the parallel training strategy will subtly harm the sampling performance, which can be observed from \cref{tab:ablation} in \cref{subsec:4.ablation}.
This is because the achieved sub-optimal $\tau_K,\cdots,\tau_{i+1}$ fail to ensure that the sampling distribution of $\widetilde{\mathbf x}_{t_i}$ matches $q_{t_i}(\mathbf x_{t_i})$.
Then $f_{\theta,\tau}(\widetilde{\mathbf x}_{t_i},\tau_i)$ and $f_{\theta,\tau}(\mathbf x_{t_i},\tau_i)$ have different distributions.
Therefore, one can only use biased $\widetilde{\mathbf x}_{t_i}$ for subsequent training of $\tau_i$.
In other words, the parallel training will introduce extra error at each denoising step.

\section{Experiments}
\label{sec:exp}

\definecolor{myred}{RGB}{213, 123, 101}
\definecolor{mygray}{RGB}{153, 153, 153}
\begin{figure*}[t]
\centering
\includegraphics[width=1.0\textwidth]{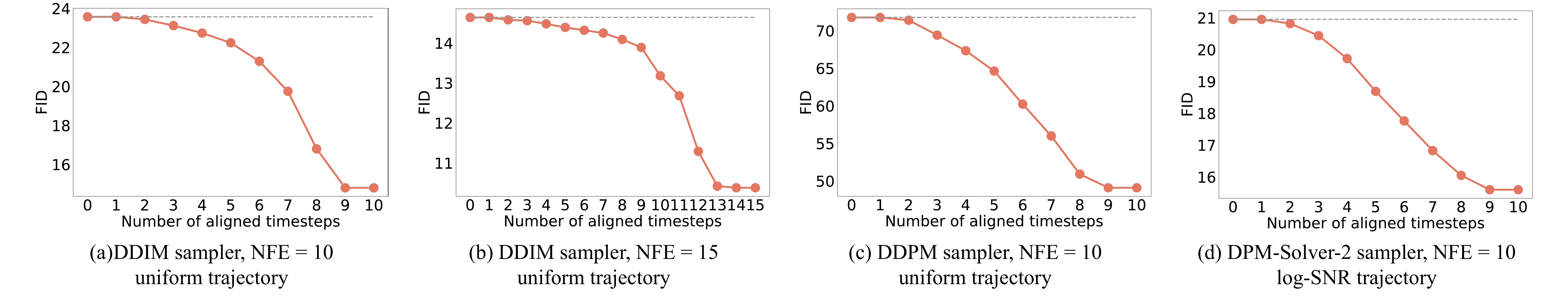}
\vspace{-20pt}
\caption{%
    \textbf{Quantitative measurement} of FID drop by DDIM, DDPM, and DPM-Solver under different trajectories.
    The horizontal axis reports the number of replaced timesteps from $t_K$ to $t_1$.
    The \textcolor{myred}{\textbf{red line}} shows the FID drop, while the \textcolor{mygray}{\textbf{gray dashed line}} shows baseline FID.
}
\label{fig:fid_step_by_step}
\vspace{-0pt}
\end{figure*}

\begin{figure*}[t]
\centering
\includegraphics[width=1.0\textwidth]{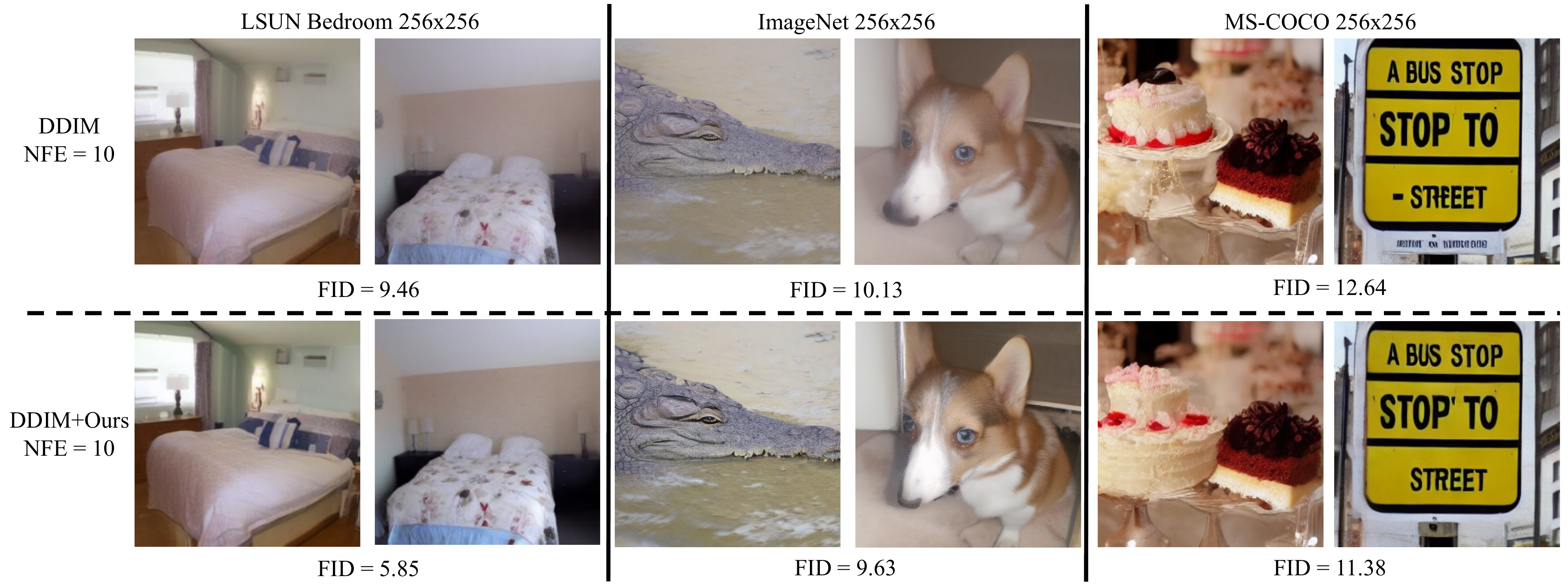}
\vspace{-20pt}
\caption{%
    \textbf{Qualitative comparison} on LSUN Bedroom, ImageNet, and MS-COCO, under LDM.
    The three tasks are unconditional, label-conditioned, and text-conditioned generation, respectively.
    All are evaluated with 10 NFEs and uniform trajectory via DDIM.
}
\label{fig:visualization}
\vspace{-5pt}
\end{figure*}

\setlength{\tabcolsep}{10pt}
\begin{table*}[!htp]
\caption{%
    \textbf{Quantitative comparison} measured by FID $\downarrow$ between \textit{sequential} and \textit{parallel} training strategies, where all are evaluated by drawing 50,000 samples via DDIM.
    We conduct experiments on CIFAR10, CelebA, LSUN Bedroom, FFHQ, and CelebA-HQ.
    We apply quadratic trajectory for CIFAR10 and CelebA datasets, and uniform trajectory on the other three datasets.
    }
\label{tab:ablation}
\vspace{-10pt}
\centering
\SetTblrInner{rowsep=0.3pt}       
\SetTblrInner{colsep=12.0pt}      
\begin{tblr}{
    cell{1-4}{1}={halign=l,valign=m},   
    cell{1-4}{2-6}={halign=c,valign=m},     
    hline{1,5}={1-6}{1.0pt},     
    hline{2}={1-6}{},          
}
Dataset                   &   CIFAR10 &    CelebA & LSUN Bedroom &      FFHQ & CelebA-HQ \\
DDIM             &     13.65 &     13.53 &         9.65 &     23.57 &     21.81 \\
DDIM + Ours (sequential) &     11.77 & \bf 11.84 &     \bf 6.07 & \bf 14.80 & \bf 17.75 \\
DDIM + Ours (parallel)   & \bf 11.75 &     12.14 &         6.08 &     15.03 &     18.03 \\
\end{tblr}
\vspace{-5pt}
\end{table*}

In this section, we show that \method can greatly improve the performance of existing DPM acceleration algorithms, varying different NFEs of calling $\boldsymbol\epsilon_\theta$.
We show great improvements of sampling quality in~\cref{subsec:4.quality}, and two different training strategies in~\cref{subsec:4.ablation}.

\subsection{Experimental Setups}

\noindent\textbf{Datasets and baselines.}
We apply \method to existing acceleration methods, including DDIM~\citep{song2020denoising}, DDPM~\citep{ho2020denoising}, Analytic-DDIM~\citep{bao2022analyticdpm}, Analytic-DDPM~\citep{bao2022analyticdpm}, and DPM-Solver-2~\citep{lu2022dpm}.
We also apply \method on EDM~\citep{Karras2022edm}, a high-order DE solver with specially designed noise schedule.
For high-resolution DPMs, we involve latent diffusion models (LDMs)~\citep{rombach2022high}.
To confirm the efficacy under extreme NFEs, we apply \method on Consistent Distillation (CD)~\citep{song2023consistency}.
%
%
The pre-trained DPMs are trained on CIFAR10~\citep{Krizhevsky_2009_17719}, CelebA~\citep{liu2015deep}, LSUN Bedroom~\citep{yu15lsun}, FFHQ~\citep{karras2019style}, CelebA-HQ~\citep{karras2018progressive}, ImageNet~\citep{deng2009imagenet}, and MS-COCO~\citep{lin2014coco}, respectively.
Note that EDM and LDM on ImageNet and MS-COCO are conditional generation, based on label and text, respectively.
DPMs on all seven datasets are under linear schedule~\citep{ho2020denoising} with $T=1,000$.
%

\noindent\textbf{Evaluation metrics.}
We draw 50,000 samples and use Fr\'{e}chet Inception Distance (FID)~\citep{heusel2017gans} to evaluate the fidelity of the synthesized images.
Inception Score (IS)~\citep{Salimans2016ImprovedTF} measures how well a model captures the full ImageNet class distribution while still producing individual samples of a single class convincingly.
To better measure spatial relationships, we involve sFID~\citep{Nash2021GeneratingIW}, rewarding image distributions with coherent high-level structure.
Finally, we use Improved Precision and Recall~\citep{Kynknniemi2019ImprovedPA} to separately measure sample fidelity (precision) and diversity (recall).

\noindent\textbf{Implementation details.}
We train \method on PyTorch~\citep{paszke2019pytorch} with NVIDIA A100 GPUs.
We use the pre-trained DDIM\footnote{https://github.com/ermongroup/ddim} of CelebA provided in the official implementation, while that on CIFAR10 is trained by Bao~\textit{et al.}~\citep{bao2022analyticdpm} with the same U-Net structure as Nichol \& Dhariwal~\citep{nichol2021improved}.
For EDM\footnote{https://github.com/NVlabs/edm}, LDM\footnote{https://github.com/CompVis/latent-diffusion}, and CD\footnote{https://github.com/openai/consistency\_models}, we directly use the pre-trained model in the official implementations.

\subsection{Sample Quality}
\label{subsec:4.quality}

\noindent\textbf{Unconditional generation on CIFAR10 and CelebA.}
For the strongest baseline, we apply the quadratic trajectory for DDIM and DDPM on both CIFAR10 and CelebA datasets, which empirically achieves better FID performance than uniform trajectory.
As for DPM-Solver-2, we use the log-SNR trajectory following the original setup~\citep{lu2022dpm}.
As shown in~\cref{fig:fid}, under all trajectories of different NFEs, our proposed \method consistently improves the sampling performance of DDIM, DDPM, Analytic-DDIM, Analytic-DDPM, and DPM-Solver-2 significantly.

\noindent\textbf{Unconditional generation on LDM.}
As for the datasets with high resolution 256x256, we apply LDM to guarantee the training efficiency of our algorithm without loss of the synthesis performance.
From~\cref{fig:fid_ldm} one can conclude that our algorithm achieves even better performance improvement on the high-resolution datasets.
To quantitatively demonstrate the improvement, we make further comparison using more metrics, as shown in~\cref{tab:metric}.
Qualitative results can be found in~\cref{fig:visualization} for clear demonstration.

\noindent\textbf{High-order sampler generation using EDM.}
As for high-order DE solver, we involve EDM which introduces the 2nd Heun sampling method on label-conditioned generation.
As demonstrated in~\cref{fig:fid_cond}, our method improves the generation performance as well, indicating the great compatibility for conditional generation under high-order DE solvers.

\noindent\textbf{Label-conditioned generation on ImageNet.}
As shown in~\cref{tab:metric} and~\cref{fig:visualization}, the synthesis performance of our algorithm surpasses the baseline qualitatively and quantitatively.

\noindent\textbf{Text-conditioned generation on MS-COCO.}
For the most challenging text-conditioned generation, qualitative and quantitative results demonstrated in~\cref{fig:fid_cond,fig:visualization} and~\cref{tab:metric} confirm the compatibility and capability of our method.

\noindent\textbf{Generation under extreme NFEs.}
Despite significant improvements on the state-of-the-art training-free acceleration methods, we also confirm the efficacy of \method on CD.
As demonstrated in~\cref{tab:metric_cd}, our method is capable of improving performance of CD under extremely small NFEs.
More comparison is addressed in \supp.

\noindent\textbf{Performance under different CFG scales.}
Classifier-Free Guidance (CFG)~\citep{ho2021classifierfree} is a commonly used technique to facilitate fidelity of conditional generation.
As in~\cref{tab:cfg}, \method is capable of consistently improving the performance on label- and text-conditioned tasks with different CFG scales.
We will address the optimized timesteps which vary for different scales in \supp.

\noindent\textbf{Comparison with other alternatives.}
From~\cref{tab:comparison} one can conclude that our method shows superior performance.
Besides, we would like to reaffirm that our method does not target finding an optimal schedule.
Instead, it works as a plug-in, bringing further improvements over baselines.

It is noteworthy that the performance improvement is more significant with a small NFE.
For instance, the improvement of FID between DDIM and DDIM + Ours on FFHQ decreases from 8.77 to 1.51 as NFE grows from 10 to 25.
This roots in the fact that larger NFE relieves the truncation error, and hence reduces the gap between the real and sampling distributions.
We will provide further analysis on optimized timesteps regarding different solvers, datasets, and trajectories in \supp.

We also confirm the performance improvement by optimizing the timesteps one by one.
As in~\cref{fig:fid_step_by_step}, replacing the timestep gradually decreases the FID monotonically, confirming the correctness of~\Cref{thm:main2} and the effectiveness of \method.
One can also verify the effectiveness from~\cref{fig:gap}.
By replacing the timestep gradually, the distribution gap consistently decreases more and more significantly, demonstrating the strong capability of correcting the one-step and hence the accumulative truncation error.

\subsection{Training Strategy of \methodtitle}\label{subsec:4.ablation}

Recall that we separately introduce a \textit{sequential strategy} and a \textit{parallel strategy} in~\cref{subsec:3.strategy}.
We deduce that there is a performance gap between them, mainly due to the extra error introduced by the parallel strategy at each denoising step.
This can also be concluded from~\cref{tab:ablation} clearly.
Nevertheless, the parallel training strategy achieves on-par sampling performance, which provides a far more efficient version of \method empirically, and is extremely significant to the training of large NFE cases.
Therefore, how to improve the performance of the parallel training strategy will be an interesting avenue for future research.

\section{Conclusion}\label{sec:conclusion}

In this paper, we propose a plug-in algorithm for more accurate DPM acceleration, which replaces the original timestep to an optimized one.
We provide a proof for an estimated error bound of the deterministic DE solver, to show the feasibility to achieve better sampling performance by simply optimizing the timestep.
We conduct comprehensive experiments to demonstrate significant improvement of sampling quality under different NFEs.

\section*{Acknowledgements}

This work was supported by Beijing Natural Science Foundation (L222008), the Natural Science Foundation of China (U2336214, 62332019), Beijing Hospitals Authority Clinical Medicine Development of special funding support (ZLRK202330), National Natural Science Foundation of China (62302297), Shanghai Sailing Program (22YF1420300), and Young Elite Scientists Sponsorship Program by CAST (2022QNRC001).

{
\small
\bibliographystyle{ieeenat_fullname}
\bibliography{ref.bib}
}

\end{document}